\documentclass[fleqn,10pt]{wlscirep}
\usepackage[utf8]{inputenc}
\usepackage[T1]{fontenc}

\usepackage{graphicx}
\usepackage{subcaption}
\usepackage{float}
\usepackage{caption}
\usepackage{hyperref}
\usepackage[export]{adjustbox}

\usepackage{authblk}        
\makeatletter
\renewcommand\@fnsymbol[1]{%
  \ifcase#1\or *\or +\or \ddagger\or \mathsection\or \mathparagraph\or \|\or **\or ++\else\@ctrerr\fi}
\makeatother

\title{A Multimodal and Multi-centric Head and Neck Cancer Dataset for Segmentation, Diagnosis and Outcome Prediction}

\author[1,*,+]{Numan Saeed}
\author[2,+]{Salma Hassan}
\author[2,+]{Shahad Hardan}
\author[1]{Ahmed Aly}
\author[2]{Darya Taratynova}
\author[1]{Umair Nawaz}
\author[1]{Ufaq Khan}
\author[1]{Muhammad Ridzuan}
\author[9,10]{Vincent Andrearczyk}
\author[9,10]{Adrien Depeursinge}
\author[1]{Yutong Xie}
\author[4]{Thomas Eugene}
\author[4]{Raphaël Metz}
\author[5]{Mélanie Dore}
\author[6]{Gregory Delpon}
\author[7]{Vijay Ram Kumar Papineni}
\author[8]{Kareem Wahid}
\author[8]{Cem Dede}
\author[8]{Alaa Mohamed Shawky Ali}
\author[8]{Carlos Sjogreen}
\author[8]{Mohamed Naser}
\author[8]{Clifton D. Fuller}
\author[9]{Valentin Oreiller}
\author[10]{Mario Jreige}
\author[10]{John O. Prior}
\author[11]{Catherine Cheze Le Rest}
\author[11]{Olena Tankyevych}
\author[12]{Pierre Decazes}
\author[12]{Su Ruan}
\author[13]{Stephanie Tanadini-Lang}
\author[14]{Martin Vallières}
\author[16,17]{Hesham Elhalawani}
\author[18]{Ronan Abgral}
\author[18]{Romain Floch}
\author[18]{Kevin Kerleguer}
\author[19]{Ulrike Schick}
\author[19]{Maelle Mauguen}
\author[18]{David Bourhis}
\author[20]{Jean-Christophe Leclere}
\author[19]{Amandine Sambourg}
\author[3,+]{Arman Rahmim}
\author[15,+]{Mathieu Hatt}
\author[1]{Mohammad Yaqub}
\affil[1]{Department of Computer Vision, Mohamed bin Zayed University of Artificial Intelligence, Abu Dhabi, UAE}
\affil[2]{Department of Machine Learning, Mohamed bin Zayed University of Artificial Intelligence, Abu Dhabi, UAE}
\affil[3]{Department of Integrative Oncology, BC Cancer Research Institute, Vancouver, BC, Canada}
\affil[4]{Nantes Université, CHU Nantes, Nuclear Medicine Department, Nantes, France}
\affil[5]{Radiation Oncology Department, Institut de Cancérologie de l’Ouest, Saint-Herblain, France}
\affil[6]{Medical Physics Department, Institut de Cancérologie de l’Ouest, Saint Herblain, France}
\affil[7]{Radiology Department, Sheikh Shakhbout Medical City, Abu Dhabi, UAE}
\affil[8]{MD Anderson Cancer Center, The University of Texas, Texas, United States}
\affil[9]{Institute of Informatics, HES-SO Valais-Wallis University of Applied Sciences and Arts, Sierre, Switzerland}
\affil[10]{Department of Nuclear Medicine and Molecular Imaging, Lausanne University Hospital (CHUV), Switzerland}
\affil[11]{Centre Hospitalier Universitaire de Poitiers (CHUP), Poitiers, France}
\affil[12]{Center Henri Becquerel, LITIS laboratory, University of Rouen Normandy, Rouen, France}
\affil[13]{University Hospital Zürich. Zurich, Switzerland}
\affil[14]{Department of Computer Science, Université de Sherbrooke, Sherbrooke, Québec, Canada}
\affil[15]{LaTIM, INSERM, UMR 1101, Univ Brest, Brest, France}
\affil[16]{Department of Radiation Oncology, Brigham and Women’s Hospital, Boston, United States}
\affil[17]{Dana Farber Cancer Institute, Harvard Medical School, Boston, USA}
\affil[18]{Nuclear Medicine Department, University Hospital of Brest, Brest, France}
\affil[19]{Radiation Oncology Department, University Hospital of Brest, Brest, France}
\affil[20]{Head and Neck Surgery Department, University Hospital of Brest, Brest, France}
\affil[*]{numan.saeed@mbzuai.ac.ae}

\affil[+]{these authors contributed equally to this work}

\begin{abstract}
We present a publicly available multimodal dataset for head and neck cancer research, comprising 1123 annotated Positron Emission Tomography/Computed Tomography (PET/CT) studies from patients with histologically confirmed disease, acquired from 10 international medical centers. All studies contain co-registered PET/CT scans with varying acquisition protocols, reflecting real-world clinical diversity from a long-term, multi-institution retrospective collection. Primary gross tumor volumes (GTVp) and involved lymph nodes (GTVn) were manually segmented by experienced radiation oncologists and radiologists following established guidelines. We provide anonymized NifTi files, expert-annotated segmentation masks, comprehensive clinical metadata, and radiotherapy dose distributions for a patient subset. The metadata include TNM staging, HPV status, demographics, long-term follow-up outcomes, survival times, censoring indicators, and treatment information. To demonstrate its utility, we benchmark three key clinical tasks: automated tumor segmentation, recurrence-free survival prediction, and HPV status classification, using state-of-the-art deep learning models like UNet, SegResNet, and multimodal prognostic frameworks.
\end{abstract}

\begin{document}

\flushbottom
\maketitle

\thispagestyle{empty}

\section*{Background and Summary}

Head and neck cancer (HNC) represents a heterogeneous group of malignancies arising from anatomically and functionally critical regions, including the oral cavity, pharynx, larynx, paranasal sinuses, and salivary glands \cite{bhat2021head}. The intricate anatomy of the head and neck, housing vital structures for speech, swallowing, and breathing, presents substantial challenges for diagnosis, treatment, and long-term disease control. Approximately 90\% of HNC cases are histologically classified as squamous cell carcinomas (HNSCC). Established risk factors include tobacco use and alcohol consumption \cite{Maso2016Combined, Dhull2018Major, Aupérin2020Epidemiology, Gormley2022Reviewing, Hashibe2009Interaction, Mody2021Head, Hashibe2007Alcohol}, which act synergistically to substantially increase disease risk (up to 35-fold in heavy users of both substances) \cite{Maso2016Combined, Hashibe2009Interaction, Maasland2014Alcohol, Zhang2015Different, Hashibe2007Alcohol, Maier1992Tobacco}.

Despite therapeutic advances, the 5-year survival rate for HNC remains low, especially in advanced or recurrent cases. Current clinical workflows, which are based on TNM (tumor size, lymph nodes, metastasis) staging, imaging, and histopathology, do not consistently enable early detection or accurate prognostication. The use of specific biomarkers, such as Human PapillomaVirus (HPV) status and gene expression profiles, has improved risk stratification in specific subgroups. Regardless, most of these biomarkers remain investigational and are not routinely integrated into clinical decision-making. These limitations underscore the need for improved diagnostic and prognostic tools capable of capturing the complex, multimodal nature of the disease.

Artificial intelligence (AI) offers a promising avenue to address these gaps by integrating diverse biomedical data sources, such as imaging (Computed Tomography (CT), Positron Emission Tomography (PET)), radiotherapy dose distributions, and clinical records (EHR). Multimodal approaches can exploit complementary information to produce more accurate, robust, and generalizable models for diagnosis, prognosis, and treatment planning \cite{Boehm2021Harnessing,Chen2022Pan-cancer,Steyaert2023Multimodal}. However, the development and rigorous validation of such models for HNC has been hindered by the scarcity of large, diverse, and publicly accessible datasets. Most existing studies rely on small, single-center cohorts, limiting reproducibility and clinical translation due to low generalizability of the trained models.

To address this issue, we present a large-scale, multi-institutional, multimodal HNC dataset collected from 10 international centers. The dataset integrates co-registered CT and PET imaging, radiotherapy dose distributions, and structured clinical data from electronic health records, complemented by expert-annotated segmentation masks, as well as diagnostic and prognostic labels. This breadth of information enables the exploration of AI methods for tumor detection and segmentation \cite{saeed2021ensemble, saeed2022tmss}, diagnostic tasks and outcome prediction \cite{saeed2024survrnc}, as well as treatment optimization. By spanning multiple institutions and a long retrospective period of time for data collection, the dataset reduces single-center bias and supports the development of models with improved generalizability across diverse populations, scanner models,  acquisition protocols and image reconstruction settings. It thus provides an unprecedented resource for advancing methodological and clinical research in HNC.

\section*{Methods}

\subsection*{Data Collection}

\begin{figure}[!t]
    \centering
    \includegraphics[width=1\linewidth]{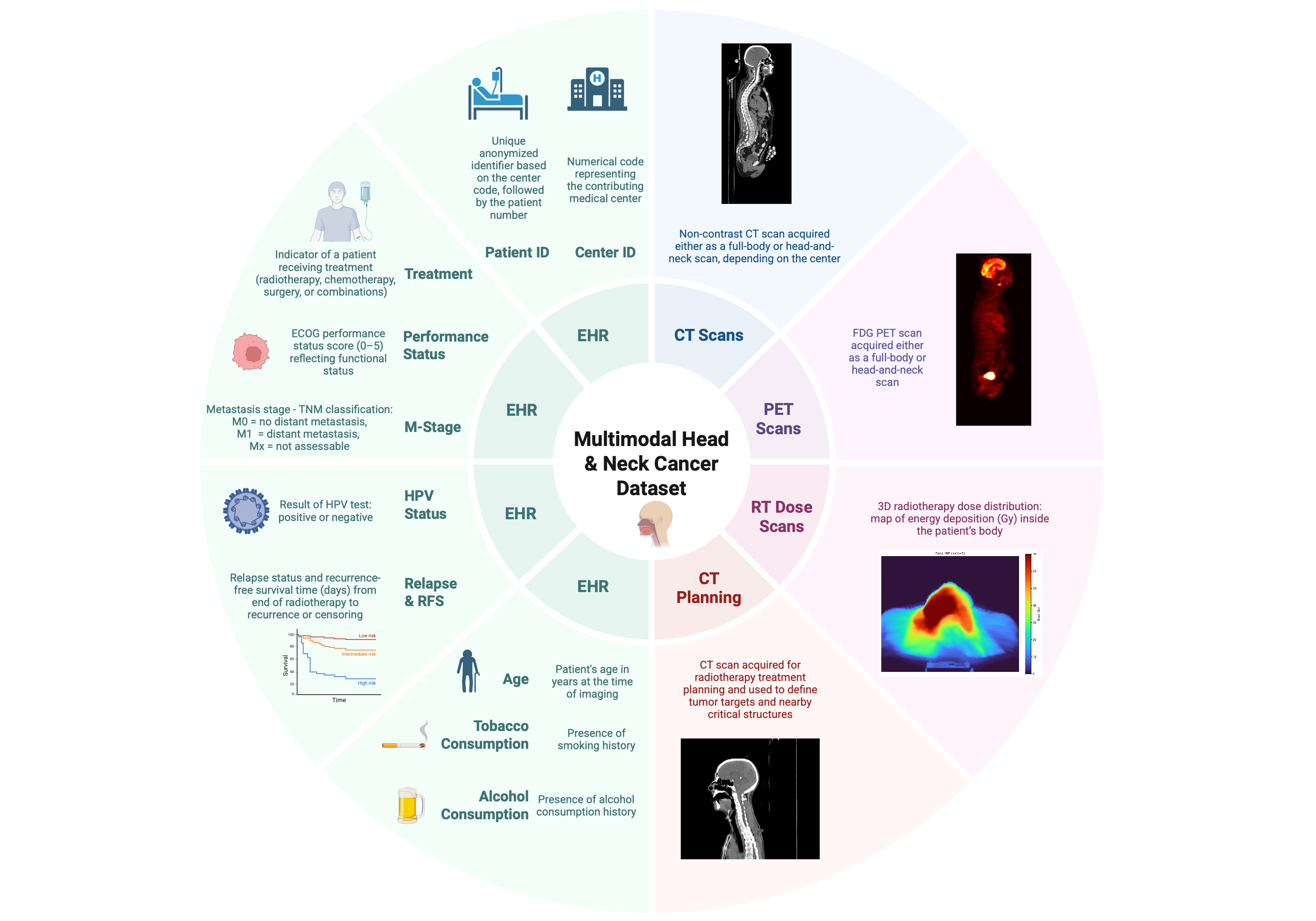}
    \caption{The dataset contains multiple modalities that aid in the diagnosis and prognosis of head and neck cancer. These modalities include CT scans, PET scans, RT dose image, CT planning images, and different clinical features.}
    \label{fig:dataset}
\end{figure}
\begin{figure*}[!t]
    \centering
    \begin{subfigure}[t]{0.49\textwidth}
        \centering
        \includegraphics[width=\textwidth]{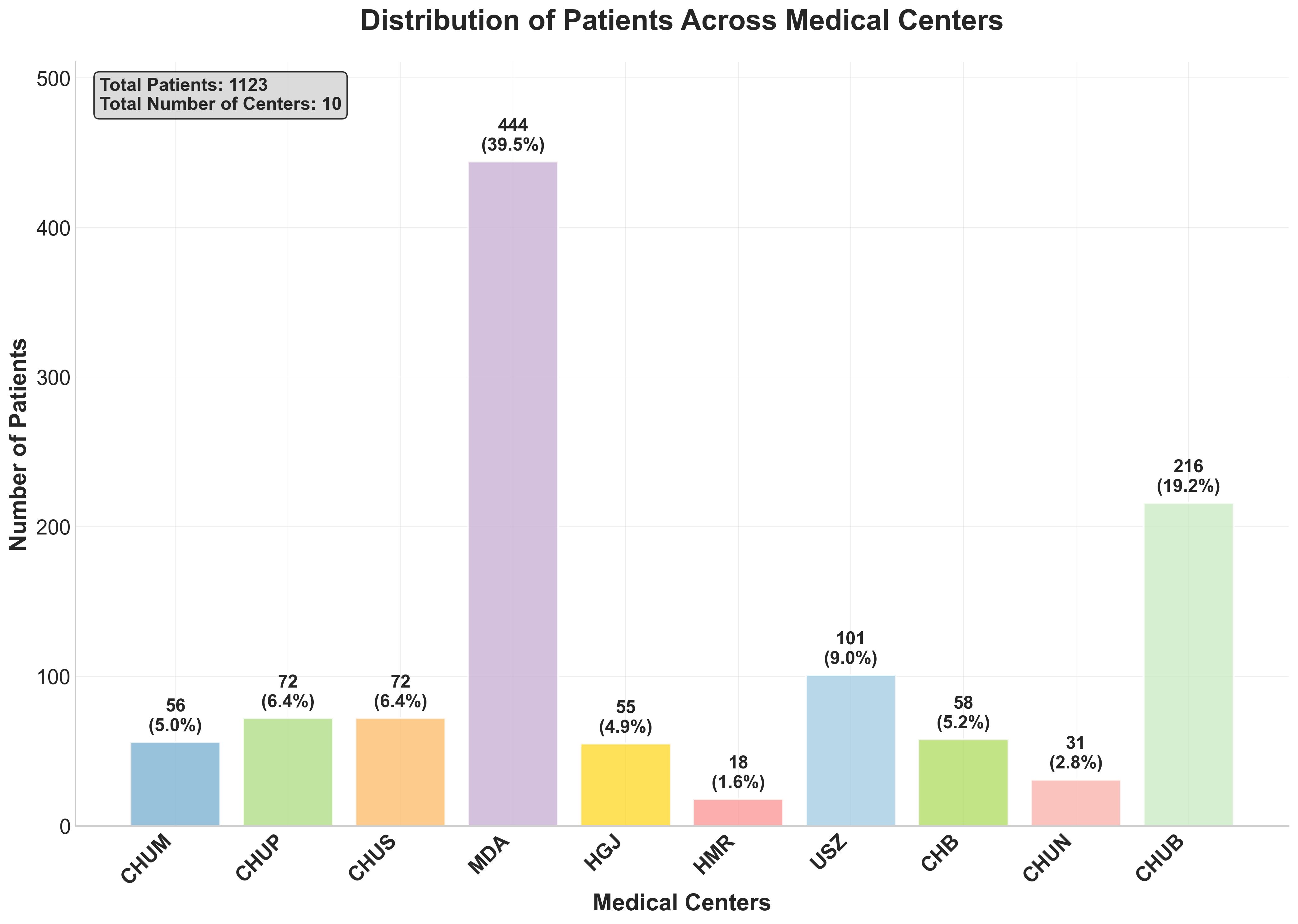}
        \caption{Distribution of 1,123 patients across ten medical centers. MDA contributes the largest proportion with 444 patients (39.5\%), followed by CHUB with 216 patients (19.2\%) and USZ with 101 patients (9.0\%).} 
        \label{fig:centers_dist}
    \end{subfigure}%
    \hfill
    \begin{subfigure}[t]{0.49\textwidth}
        \centering
        \includegraphics[width=\textwidth]{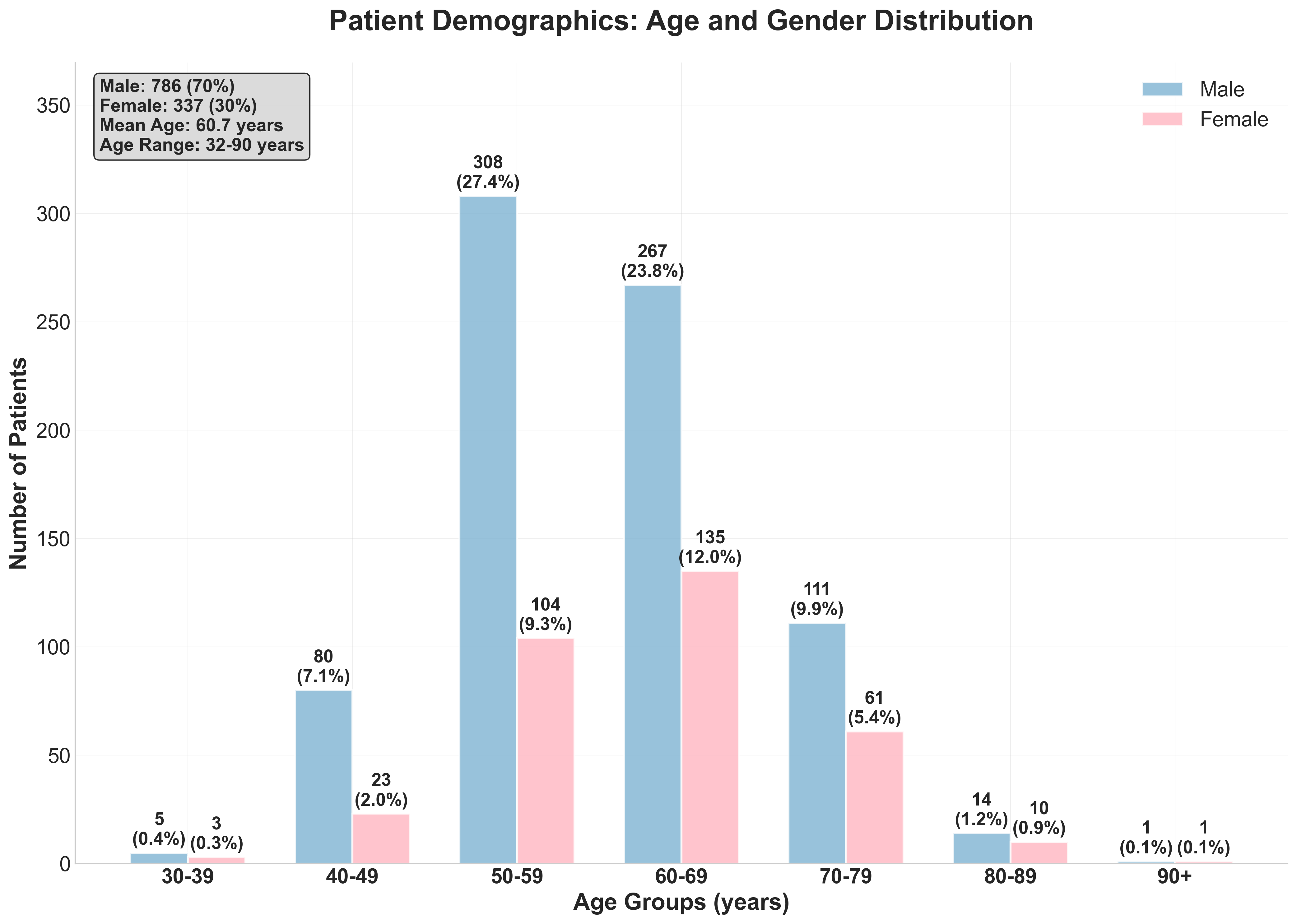}
        \caption{Patient demographics showing age and gender distribution. The cohort demonstrates male predominance (70\%) across all age groups, with peak representation in the 50-69 years range (72.5\% of patients). Mean age is 60.5 years (range: 32-90 years).}
        \label{fig:demographics_dist}
    \end{subfigure}
    \caption{Patient characteristics and distribution across the multi-centric cohort.}
    \label{fig:patient_characteristics}
\end{figure*}

The dataset is a large-scale, multi-institutional collection of multimodal PET/CT scans and rich clinical data from 1,123 patients with histologically confirmed HNC. Fig. \ref{fig:dataset} shows the variety of modalities available in the dataset. As illustrated in Fig. \ref{fig:centers_dist}, the data is sourced from 10 international medical centers, and captures diversity in scanner models, acquisition and reconstruction protocols, and patient demographics, offering a realistic foundation for developing models that may generalize better across different institutions. The participating clinical centers, along with their acronyms, country, and PET/CT scanner models, are summarized in Table \ref{tab:center_info}. It is one of the largest publicly available resources of its kind, featuring standardized three-dimensional imaging, expert-delineated segmentation, and curated metadata that include staging, treatment intent, and long-term follow-up outcomes. This breadth and depth make it a powerful benchmark for image-based prediction and for creating models with real clinical value.

The dataset has been designed to support three complementary tasks. The first is segmentation of the primary gross tumor volume (GTVp) and involved lymph nodes (GTVn). These segmentation masks were generated manually by experienced radiation oncologists or radiologists, following institutional pre-established guidelines and undergoing quality checks before release. They are co-registered with the PET/CT scans. The second task is recurrence-free survival (RFS) prediction, combining imaging and clinical variables with time-to-event data, enabling robust prognostic modeling. The third task is diagnosis of HPV status, allowing exploration of imaging and clinical correlates of this important biomarker. 

The cohort composition reflects its multi-centric nature. MD Anderson center (MDA) is the largest contributor, with 444 patients (39.6\%), followed by Centre Hospitalier Universitaire de Brest (CHUB) with 216 patients (19.2\%) and the University Hospital of Zurich (USZ) with 101 patients (9.0\%). The remaining seven centers contribute between 18 and 72 patients each, as shown in Fig. \ref{fig:centers_dist}. This broad institutional representation provides the variation in imaging and patient populations that is critical for testing model generalization.

\begin{figure*}[!t]
    \centering
    \includegraphics[width=0.85\textwidth]{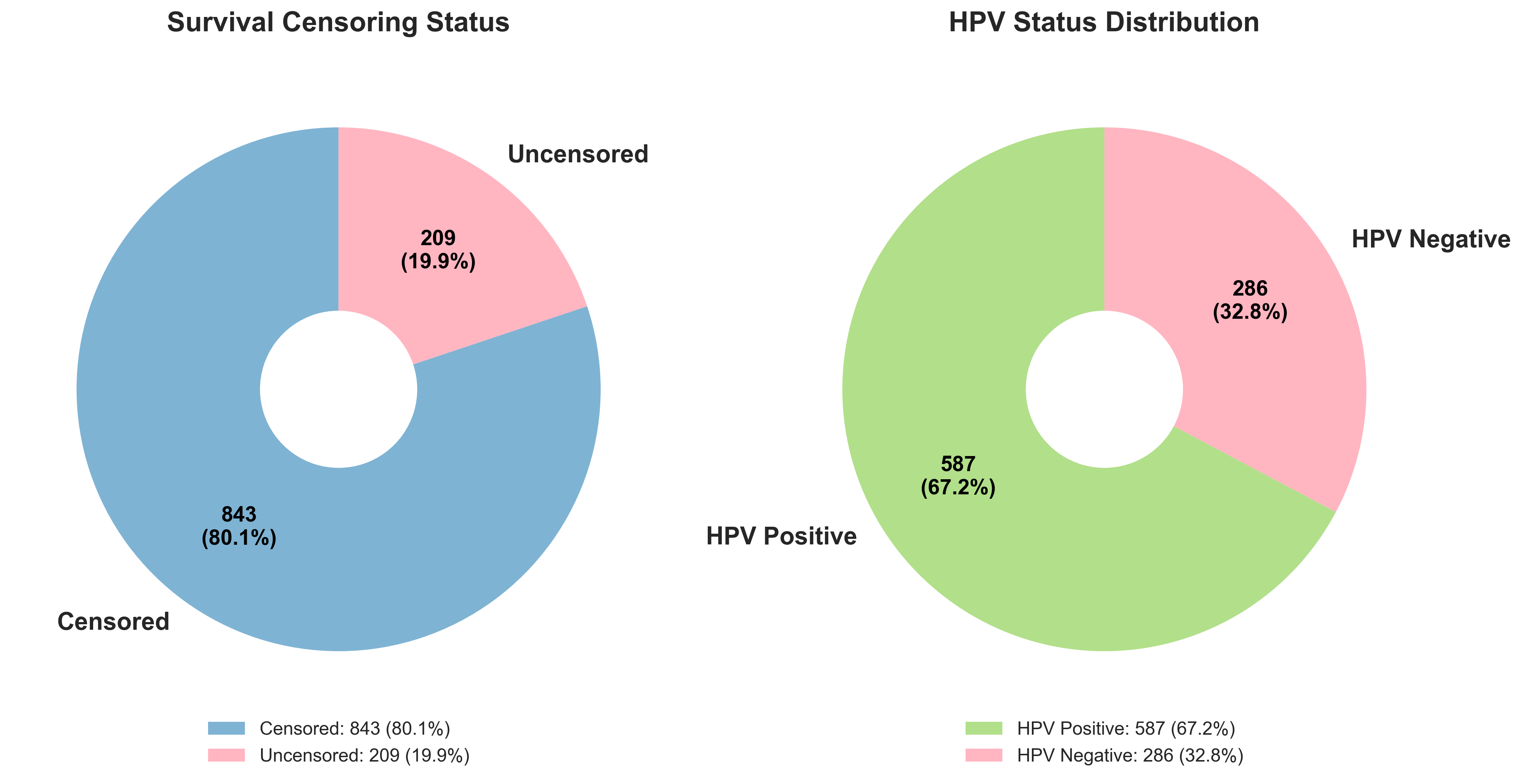}
    \caption{Clinical characteristics in dataset. Left: Survival censoring status showing the distribution of censored (843 patients, 80.1\%) and uncensored (209 patients, 19.9\%) cases among patients with available follow-up data (n=1,052). The high censoring rate indicates good long-term follow-up with most patients event-free at last contact. Right: HPV status distribution among patients with known HPV status (n=873), showing HPV-positive (587 patients, 67.2\%) and HPV-negative (286 patients, 32.8\%) cases . The HPV-positive predominance is consistent with typical HNC epidemiology.}
    \label{fig:clinical_characteristics}
\end{figure*}

Clinical data were harmonized across all centers to include RFS times, censoring indicators, HPV status, demographics, and tumor staging. Fig. \ref{fig:clinical_characteristics} shows that 843 patients (80.1\%) were censored, while 209 (19.9\%) experienced a recurrence or death during follow-up. Regarding the 873 patients for which the HPV status information is available,  587 patients (67.2\%) were positive and 286 (32.8\%) were negative( \textbf{Fig. \ref{fig:clinical_characteristics}}). The age distribution spans from 21 to 92 years, with a median of 60 years, and reveals clear demographic trends (Fig. \ref{fig:demographics_dist}). The majority of patients are concentrated in the 50–69-year age range, accounting for more than 70\% of the dataset, consistent with the peak incidence period for HNC. Males predominate in nearly every age bracket, particularly within this peak range, where the male-to-female ratio is most pronounced. This imbalance reflects established epidemiological patterns, influenced by historically higher exposure to risk factors such as tobacco use, alcohol consumption, and certain occupational hazards among men.

\begin{table}[ht]
\centering
\caption{Overview of participating centers, their country of origin, acronyms, and the PET/CT scanners used in the study}
\label{tab:center_info}
\begin{tabular}{@{}p{7cm}p{2cm}p{2cm}p{5cm}@{}}
\toprule
\textbf{Center} & \textbf{Country} & \textbf{Acronym} & \textbf{PET/CT scanner} \\ \midrule
Hôpital général juif & Canada & HGJ & Discovery ST, GE Healthcare \\
Centre hospitalier universitaire de Sherbrooke & Canada & CHUS & GeminiGXL 16, Philips \\
Hôpital Maisonneuve-Rosemont & Canada & HMR & Discovery STE, GE Healthcare \\
Centre hospitalier de l’Université de Montréal & Canada & CHUM & Discovery STE, GE Healthcare \\
Centre Hospitalier Universitaire de Poitiers & France & CHUP & Biograph mCT 40 ToF, Siemens \\
MD Anderson Cancer Center & USA & MDA & Discovery HR, Discovery RX, Discovery ST, Discovery STE (GE Healthcare) \\
University Hospital of Zürich & Switzerland & USZ & Discovery HR, Discovery RX, Discovery STE, Discovery LS, Discovery 690 (GE Healthcare) \\
Centre Henri Becquerel & France & CHB & GE710, GE Healthcare \\
Centre Hospitalier Universitaire de Brest & France & CHUB & Philips GEMINI, Siemens Biograph, Siemens Biograph Vision \\
Centre Hospitalier Universitaire de Nantes & France & CHUN & Siemens mCT 64 vision \\ \bottomrule
\end{tabular}
\end{table}



\subsection*{Acquisition Settings and Data properties}
Table \ref{tab:pet_ct_params} summarizes the PET/CT acquisition parameters for all participating centers, including injected dose, uptake time, scanner model(s), acquisition duration, image resolution, and CT settings. Values are presented as medians with ranges where available to highlight inter-center variability in imaging protocols.

\begin{table}[!b]
\centering
\caption{PET/CT acquisition parameters across centers. Values are median (range) unless stated otherwise. Var.  indicates that the value is not fixed and may differ across cases.}
\label{tab:pet_ct_params}
\resizebox{\textwidth}{!}{%
\begin{tabular}{lcccccccc}
\hline
\textbf{Center} & \textbf{PET Dose (MBq)} & \textbf{Uptake (min)} & \textbf{PET System} & \textbf{PET Time/Pos (s)} & \textbf{PET Slice (mm)} & \textbf{PET In-plane ($mm^2$)} & \textbf{CT kVp} & \textbf{CT Slice (mm)}\\
\hline
HGJ  & 584 (368–715)        & 90     & Discovery ST (GE)                  & 300 (180–420)   & 3.27                 & 3.52×3.52 (3.52–4.69)           & 140                 & 3.75 \\
CHUS & 325 (165–517)        & 90     & Gemini GXL 16 (Philips)            & 150 (120–151)   & 4                    & 4×4                             & 140 (12–140)        & 3 (2–5) \\
HMR  & 475 (227–859)        & 90     & Discovery STE (GE)                 & 360 (120–360)   & 3.27                 & 3.52×3.52 (3.52–5.47)           & 140 (120–140)       & 3.75 \\
CHUM & 315 (199–3182)       & 90     & Discovery STE (GE)                 & 300 (120–420)   & 4 (3.27–4)           & 4×4 (3.52–5.47)                 & 120 (120–140)       & 1.5 (1.5–3.75) \\
CHUP & 421$\pm$98 (220–695) & 60$\pm$5 & Biograph mCT 40 ToF (Siemens)    & 150             & 5                    & 4×4×4 $mm^3$                     & 120                 & 5 \\
MDA  & 401 (327–266)        & 90     & Multiple hybrid PET/CT             & 180 (90–300)    & 3.27 (2.99–5)        & 5.46×5.46 (2.73–5.46)           & 120 (100–140)       & 3.75 (2.99–5) \\
USZ  & 178–513              & $\sim$60 & Multiple hybrid PET/CT           & Var.            & Var.                 & Var.                             & Var.                & Var. \\
CHB  & $\sim$3 MBq/kg       & 90$\pm$5 & GE710 (GE)                        & Var.            & 3.27                 & 2.73×2.73                        & 120                 & 2.5 \\
CHUN & 204 (129–354)        & 60     & Biograph mCT 60/40 (Siemens)       & Var.            & 2                    & 4×4                              & 120                 & 2.0 \\
CHUB & $\sim$6 or 3 MBq/kg       & 60     & Philips Gemini G-XL 6 , Siemens Biograph mCT and Siemens Biograph Vision600         & Var.            & 4, 2 and 1.6                    & 4×4, 4×4 and 1.6×1.6                          & 120                 & 4.0, 3.0 and 2.0 \\
\hline
\end{tabular}}

\end{table}

\subsection*{Annotation Protocol}
\textbf{CHUS, HGJ, and HMR:} The GTVp and GTVn contours were delineated by an experienced radiation oncologist using a radiotherapy treatment planning system. In 40\% of the training set (80 cases), the contours were created directly on the CT component of the PET/CT scan and then applied for treatment planning. The remaining 60\% (121 cases) were contoured on a separate CT scan acquired specifically for treatment planning, and subsequently aligned to the FDG-PET/CT reference frame via intensity-based free-form deformable registration in MIM software (MIM Software Inc., Cleveland, OH). The number of annotators involved in the training set is not documented.

\textbf{CHUP:} The primary tumor’s metabolic volume was first obtained automatically using the Fuzzy Locally Adaptive Bayesian (FLAB) method dedicated to PET tumor volume delineation \cite{hatt2009fuzzy}. These results were then reviewed and manually refined by a single expert with reference to the CT images, making adjustments in cases where the PET-based segmentation encompassed air or non-tumor tissues visible on the CT.

\textbf{MDA:} Contours were applied on the CT image, while co-registered PET images were used to aid physicians in visualizing the tumor. Then, these contours were updated according to the guidelines mentioned in \textbf{Annotation Guidelines}.

\textbf{USZ:} The primary tumor was delineated independently on the CT and PET scans. CT-based segmentation was carried out manually, with two radiation oncologists, each with over 10 years of experience, participating in the process. The resulting contours were then refined to account for metal artifacts, ensuring exclusion of any non-tumor-related effects. If artifacts were present in a given tumor slice, the contour for that slice was completely removed. Tumors with more than 50\% of their volume deemed unsuitable for analysis were excluded from the study. Furthermore, voxels falling outside the soft tissue Hounsfield unit (HU) range of 20 to 180 HU were eliminated. For the PET scans, tumor segmentation was performed automatically using a gradient-based approach available in MIMVISTA (MIM Software Inc., Cleveland, OH).

\textbf{CHB:} A senior nuclear medicine physician manually drew the contours for each patient for both the GTVp and GTVn using the software PET VCAR (GE Healthcare) on each FDG-PET/CT, using adaptive thresholding with visual control using merged PET and CT information. 

\textbf{CHUN:} The CHUN contours were created using an nn-UNET automatic segmentation approach using the full training and testing sets of the HECKTOR 2022 data, with careful post-processing of results to ensure higher accuracy. Later, a consultant physician in a clinical imaging department reviewed the contours in a case-by-case manner, following the guidelines in the \textbf{Annotation Guidelines} section. This review process confirmed the contours of the GTVp and GTVn for correctly segmented cases and adjusted the cases missed by the model. After the initial CT/PET, a subset of confirmed cases had a follow-up CT with RTDose. The follow-up CT was registered to the initial CT, and the resulting transformation matrix was applied to the RTDose to bring it into the same space as the initial CT/PET.

\subsubsection*{Annotation Guidelines}
Experts conducted quality control across both the training and test datasets to maintain consistency in the definition of ground-truth contours. When needed, they re-annotated the data to reflect the actual tumor volume, which was often smaller than the volumes originally outlined for radiotherapy. The contouring process was centralized through a shared cloud platform (MIM Cloud Software Inc.) to standardize the annotation environment. For cases lacking original radiotherapy GTVp or GTVn contours, experts generated annotations using PET/CT fusion in conjunction with N staging information. A set of guidelines for this quality control process was established by the expert panel, as detailed below. Cases showing PET–CT misregistration were excluded.

\textbf{Guidelines for primary tumor annotation in PET/CT images:} Oropharyngeal lesions are delineated on PET/CT by integrating information from both PET and unenhanced CT scans. The contour should encompass the complete boundaries of the morphological abnormality seen on unenhanced CT, typically visualized as a mass effect, along with the corresponding hypermetabolic volume identified on PET. This is achieved using PET, unenhanced CT, and PET/CT fusion views obtained through automatic co-registration. Areas of hypermetabolic uptake extending beyond the physical borders of the lesion, such as into the airway lumen or over bony structures without morphological evidence of invasion, are excluded. The standardized naming convention follows AAPM TG-263, with the primary tumor labeled as GTVp.
The special considerations in this regard are to verify the clinical nodal category to ensure exclusion of adjacent FDG-avid and/or enlarged lymph nodes (e.g., submandibular, high level II, or retropharyngeal nodes). For cases showing fullness or enlargement of the tonsillar fossa or base of tongue without corresponding FDG uptake, consult the clinical datasheet to rule out pre-radiation tonsillectomy or extensive biopsy. Such cases should be excluded if confirmed.

\textbf{Guidelines for nodal metastases tumor annotation in PET/CT images:} Lymph node contours are defined on PET/CT by integrating data from both PET and unenhanced CT scans. Delineation should encompass the full boundaries of the morphologically abnormal lymph node observed on unenhanced CT, along with the corresponding hypermetabolic region identified on PET. This process uses PET, unenhanced CT, and PET/CT fusion views obtained through automatic co-registration, covering all cervical lymph node levels. Inclusion criteria for metastatic nodes are: standardized uptake value (SUV) greater than 2.5, or a short-axis diameter of at least 1 cm, or pathological confirmation. The standardized label for lymph node regions of interest is GTVn. Areas of hypermetabolic uptake extending beyond the anatomical limits of the node, such as into adjacent bony, muscular, or vascular structures, are excluded from the contour. If multiple nodes are touching or merging, they should be contoured as a single structure. GTVn volumes must always be kept separate from GTVp volumes.

\subsection*{Annotation Tool} \label{sec:annotation_tool}
To ensure accuracy and consistency of tumor delineations, a dedicated annotation tool was developed for use by board-certified radiation oncologists and radiologists. The tool integrates PET/CT viewing, segmentation, and quality control within a single interface, enabling precise identification and labeling of GTVp and GTVn according to standardized protocols.

The interface combines axial PET/CT fusion visualization with adjustable CT windowing and opacity controls, allowing annotators to optimize image contrast and highlight metabolic or structural features. Layer controls manage segmentation transparency, label selection, and editing modes, while patient navigation and progress tracking ensure a systematic annotation workflow. Segmentation follows embedded protocols defining inclusion criteria for GTVp (entire morphological anomaly on CT plus corresponding hyper-metabolic PET regions, excluding non-tumor uptake) and GTVn (morphologically abnormal lymph nodes with size thresholds). Workflow features include smart interpolation between slices, manual refinement tools, cleanup functions to remove stray voxels, and mandatory separation of contiguous GTVp/GTVn regions. This standardized platform reduces inter-observer variability and ensures high-quality, reproducible annotations for robust benchmarking in HNC AI research.

\section*{Data Records}
\subsection*{Format}
\begin{figure}[!h]
    \centering
    \includegraphics[width=0.7\linewidth]{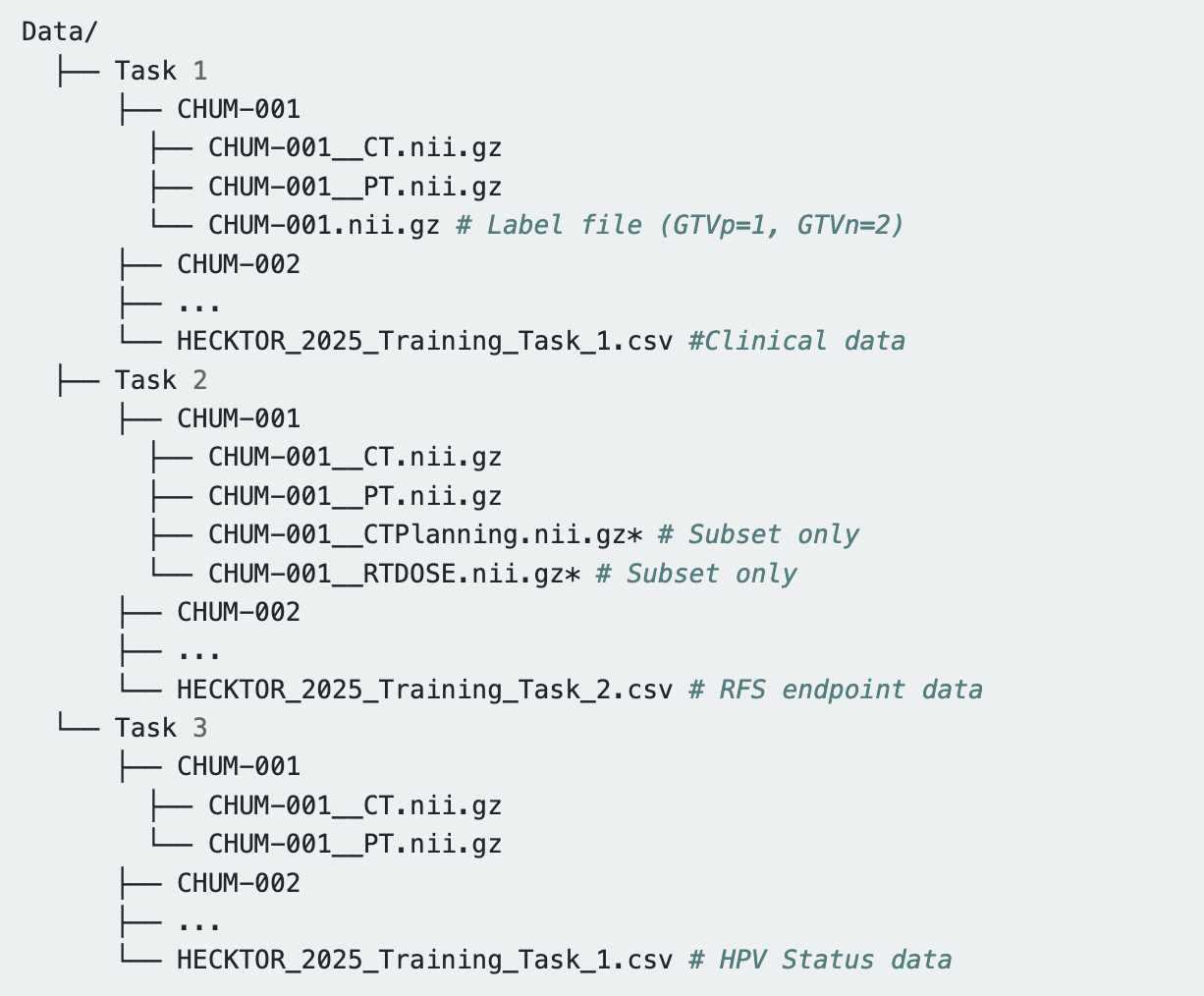}
    \caption{Dataset structure. Patients are identified by a unique, anonymized ID, and all studies of a single patient are stored under the respective patient folder. Data is organized into a separate subfolder for each task.}
    \label{fig:data_structure}
\end{figure}

The dataset is organized by task, with separate subfolders for each of the three clinical tasks, as illustrated in Fig. \ref{fig:data_structure}. Within each task directory, patient-level folders contain co-registered PET and CT scans in NIfTI format. In addition, expert-generated segmentation masks for GTVp and GTVn are also included. For Task 2, a subset of patients includes additional modalities, namely dosimetry planning CT, and radiotherapy dose map, provided in the same format and aligned with the imaging volumes. Clinical metadata is provided in .csv files, with structured fields for patient ID, center, age, alcohol consumption, TNM staging, HPV status, performance status, recurrence status, and RFS in months. This task-based folder structure, combined with standardized file formats and naming conventions, facilitates modular loading and integration into automated deep learning pipelines. The details regarding the available data fields are shown in Fig. \ref{fig:dataset}.

\subsection*{Dataset Limitations}
A first limitation is that contours were generated using PET/(unenhanced) CT fusion, whereas modalities like contrast-enhanced CT are better suited for defining accurate contours in radiation oncology. Nevertheless, given that the intended clinical use is radiomics, this level of accuracy is less critical than it would be for radiotherapy planning. Another limitation stems from variability in the ground truth annotations. Even with established guidelines and quality control measures, differences in annotation practices persisted. For example, in the USZ test set only, lesions with metal artifacts were excluded, and in some cases, re-segmentation was performed within a specific HU range. Variations also arose from differences in the annotators’ expertise and profiles. Despite attempts to standardize the contours, these inconsistencies introduced a notable amount of noise into the labeled data used for training. Additionally, there were challenges faced related to the availability of annotators to work on the 216 cases coming from the CHUB center. This disables leveraging this considerable number of scans for any segmentation tasks where ground truth is needed. A third limitation concerns the HPV status, as the information was not available for a number of patients (n=250), for the most part because at the time of diagnosis and management for these patients, it was not yet standard clinical practice to analyze HPV status.

\begin{table}[ht]
\centering
\caption{Segmentation Results on GTVp and GTVn for different models. Best values are in \textbf{bold}.}
\begin{tabular}{lcccc}
\hline
\textbf{Model} & \textbf{GTVp\_mean\_dice} & \textbf{GTVn\_DSC\_agg} & \textbf{GTVn\_F1\_agg} \\
\hline
UNet & 0.62 $\pm$ 0.03 & 0.70 $\pm$ 0.01 & \textbf{0.32 $\pm$ 0.09} \\
SegResNet & \textbf{0.67 $\pm$ 0.02} & \textbf{0.73 $\pm$ 0.01 }& 0.54 $\pm$ 0.04 \\
SwinUNETR & 0.64 $\pm$ 0.04 & 0.72 $\pm$ 0.01 & 0.15 $\pm$ 0.05 \\
\hline
\end{tabular}
\label{tab:segmentation_results}
\end{table}

\section*{Technical Validation}
To demonstrate the potential usefulness and validity of our dataset, we performed three representative downstream tasks relevant to HNC research: (1) automated tumor segmentation, (2) patient prognosis prediction, and (3) HPV status classification. For each task, we trained and evaluated state of the art machine learning models to establish performance benchmarks and showcase the dataset's potential.

\begin{figure}
    \centering
    \includegraphics[width=1\linewidth]{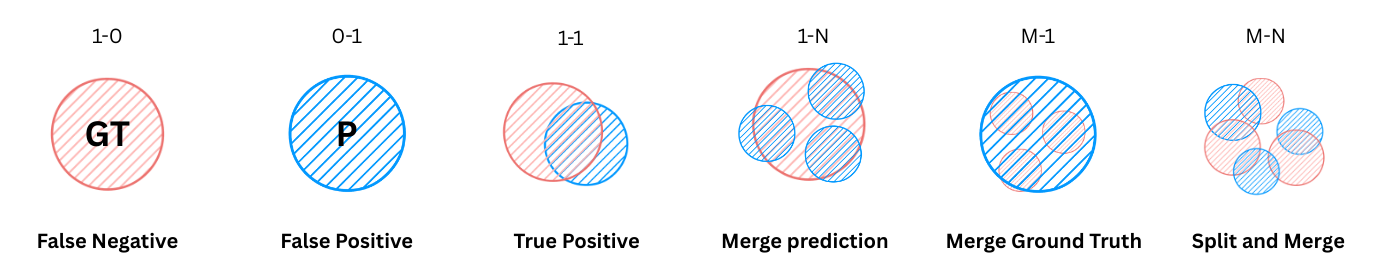}
    \caption{Illustration of prediction–ground truth relationships in nodule segmentation. Groups of predictions and ground truths are fused, and counted as a True Positive (TP) if the composite IoU $\geq$ 30$\%$. Otherwise, all predictions in the group are counted as False Positives (FP) and all ground truths as False Negatives (FN).}
    \label{fig:f1score}
\end{figure}
\subsection*{Deep Learning-based Tumor Segmentation}
To showcase the dataset for segmentation tasks, we evaluated three deep learning architectures: UNet, SegResNet, and SwinUNETR. For model input, CT and PET images were preprocessed by 1) resampling to a $1 \times 1 \times 1 mm^3$ isotropic resolution, 2) cropping to a uniform head and neck region of interest, and 3) normalizing intensity values. The processed images were then concatenated to form a 2-channel input. Using a 5-fold cross-validation scheme (80/20 split), we evaluated the models' ability to segment primary tumors and nodal metastases.

We implemented three complementary metrics for segmentation performance evaluation. The Dice Similarity Coefficient (DSC) is used to quantify the volumetric overlap of the primary tumor. However, for lymph node segmentation, the DSC can be unreliable in cases without ground-truth volumes. In such cases, even a single false negative results in a DSC of zero. To mitigate this, the aggregated Dice Similarity Coefficient ($\text{DSC}_{\text{agg}}$) \cite{HECKTOR2022} is used to provide a more stable estimate by aggregating overlap across all cases, as shown in Equation (1), where the summation is taken over all cases. To also capture detection performance, we employ the F1-score at the lesion level \cite{carass2020evaluating}. The matching rules for these cases, including merges and splits, are illustrated in Figure~\ref{fig:f1score}. Together, these metrics provide a comprehensive evaluation that captures not only spatial overlap but also detection sensitivity.

\begin{equation}
\begin{aligned}
\quad \quad \quad 
\mathrm{DSC} &= \frac{2 |P \cap GT|}{|P| + |GT|}, \quad
\mathrm{DSC}_{\mathrm{agg}} &= \frac{2 \sum_i |P_i \cap GT_i|}{\sum_i |P_i| + \sum_i |GT_i|}, \quad
F1 &= \frac{2 \cdot TP}{2 \cdot TP + FP + FN}.
\end{aligned}
\end{equation}

\begin{figure}
    \centering
    \includegraphics[width=1\linewidth]{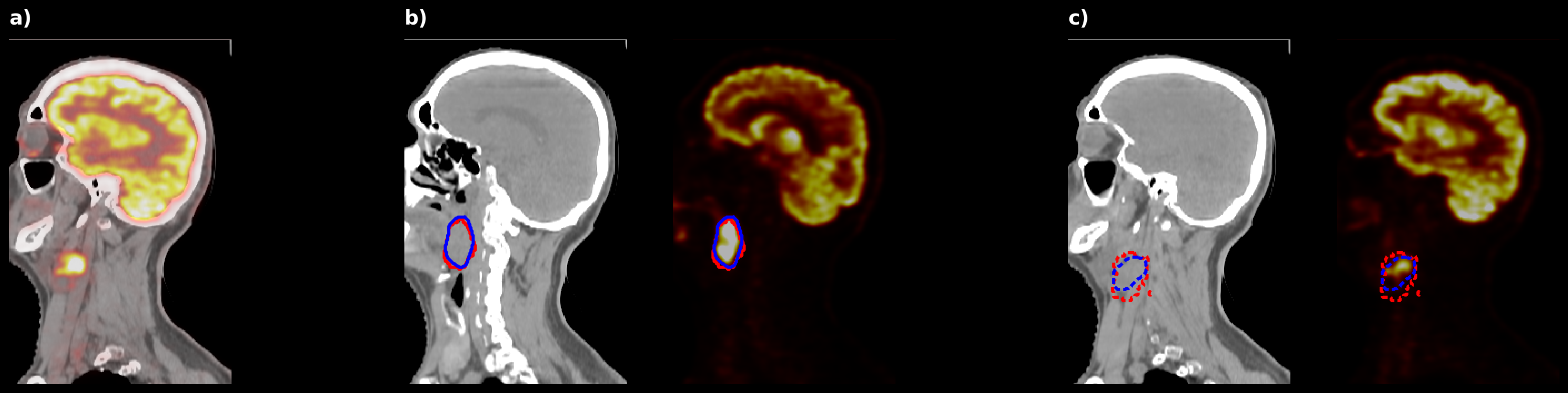}
    \caption{Segmentation results from the SegResNet model. (a) Preprocessed sagittal CT slice with PET overlay after head–neck cropping. (b) Primary tumor slice with ground truth (red) and prediction (blue) contours (solid). (c) Lymph node slice from a different location with ground truth and prediction contours (dashed).}
    \label{fig:segmentation_results}
\end{figure}

As detailed in Table \ref{tab:segmentation_results}, the SegResNet model achieved the best overall performance, with a mean DSC of $0.6660 \pm 0.0199$ for primary tumors and an aggregated DSC of $0.7291 \pm 0.0092$ for nodal metastases. Figure~\ref{fig:segmentation_results} provides qualitative visualization. These results confirm the dataset's high quality and suitability for training and validating robust segmentation algorithms. 

\begin{table}[!t]
\centering
\caption{C-index and Brier scores for different modalities and models. Best values per column within each modality group are in \textbf{bold}. Only 187 patients were available with RTDose compared to 680 for other modalities.}
\label{tab:task2_prognosis}
\begin{tabular}{llcc}
\toprule
\textbf{Modality} & \textbf{Model} & \textbf{C-Index ($\uparrow$) } & \textbf{Brier Score ($\downarrow$)} \\
\midrule
EHR & CoxPH     & 0.59 $\pm$ 0.06 & \textbf{0.16 $\pm$ 0.04} \\
EHR & DeepHit   & 0.53 $\pm$ 0.06 & 0.19 $\pm$ 0.02 \\
EHR & MTLR      & 0.55 $\pm$ 0.05 & 0.17 $\pm$ 0.02 \\
EHR & ICARE     & \textbf{0.60 $\pm$ 0.06} & 0.26 $\pm$ 0.01 \\
\midrule
EHR + CT + PET & CoxPH   & 0.61 $\pm$ 0.06 & 0.26 $\pm$ 0.01 \\
EHR + CT + PET & DeepHit & 0.55 $\pm$ 0.07 & 0.21 $\pm$ 0.02 \\
EHR + CT + PET & MTLR    & 0.59 $\pm$ 0.09 & \textbf{0.17 $\pm$ 0.01} \\
EHR + CT + PET & ICARE   & \textbf{0.67 $\pm$ 0.06} & 0.23 $\pm$ 0.01 \\
\midrule
EHR + CT + PET + RTDose & CoxPH   & 0.59 $\pm$ 0.08 & 0.26 $\pm$ 0.02 \\
EHR + CT + PET + RTDose & DeepHit & 0.53 $\pm$ 0.08 & 0.22 $\pm$ 0.02 \\
EHR + CT + PET + RTDose & MTLR    & 0.51 $\pm$ 0.08 & \textbf{0.17 $\pm$ 0.02} \\
EHR + CT + PET + RTDose & ICARE   & \textbf{0.61 $\pm$ 0.08} & 0.24 $\pm$ 0.02 \\
\bottomrule
\end{tabular}
\end{table}

\subsection*{Deep Learning-based Prognosis}

To demonstrate the dataset's utility for clinical outcome modeling, we trained models to predict RFS. We benchmarked four prognostic models (CoxPH \cite{cox1972regression}, DeepHit \cite{lee2018deephit}, MTLR \cite{yu2011learning}, and ICARE \cite{rebaud2022simplicity}) using different data combinations: Electronic Health Records (EHR) alone, EHR combined with CT and PET imaging-based radiomics features, and a subset including radiotherapy dose (RTDose) maps. Model performance was evaluated using the Concordance Index (C-index) for discrimination (higher is better) and the Brier score for calibration (lower is better).

As detailed in Table~\ref{tab:task2_prognosis}, integrating imaging with EHR data consistently improved prognostic performance. The ICARE model achieved the highest discriminative power when trained on EHR, CT, and PET data, yielding a C-index of $0.6705 \pm 0.0608$. The MTLR model showed the best calibration on the same data combination, with the lowest Brier score of $0.1653 \pm 0.0130$. While the addition of RTDose maps was explored, it led to mixed results with some models benefiting from dosimetric information while others experienced slight performance decreases, likely attributable to the substantially reduced sample size (187 versus 680 patients) available for this modality, highlighting the trade-off between data richness and statistical power. Overall, these benchmarks validate the dataset's value for developing and comparing sophisticated, multimodal survival models.

\begin{table}[t]
\centering
\caption{Classification Results on 5 different model settings using CT, PET, and EHR data. Best values are in \textbf{bold}.}
\begin{tabular}{lcccccc}
\hline
\textbf{Model} & \textbf{F1-Macro} & \textbf{AUC} & \textbf{Sensitivity} & \textbf{Specificity} & \textbf{Balanced Accuracy} & \textbf{F1-weighted} \\
\hline
DenseNet121 + MLP & 0.64 $\pm$ 0.13 & 0.77 $\pm$ 0.22& 0.99$\pm$ 0.04 & 0.22  $\pm$ 0.1918& 0.61 $\pm$ 0.0968& 0.89 $\pm$ 0.0360\\
ResNet18 + MLP    & \textbf{0.77 $\pm$ 0.09} & \textbf{0.94 $\pm$ 0.03} & 0.99 $\pm$ 0.00 & \textbf{0.43 $\pm$ 0.15}    & \textbf{0.71 $\pm$ 0.07} & \textbf{0.93 $\pm$ 0.02} \\
ViT + MLP & 0.52 $\pm$ 0.11 & 0.852 $\pm$ 0.05& 1.00  $\pm$ 0.00 & 0.07 $\pm$ 0.15& 0.53 $\pm$ 0.07& 0.86 $\pm$ 0.03 \\
EfficientNet + MLP & 0.55 $\pm$ 0.11 & 0.58 $\pm$ 0.25 & 0.99 $\pm$ 0.03& 0.12 $\pm$ 0.16& 0.55 $\pm$ 0.07 & 0.87 $\pm$ 0.03 \\
SENet + MLP & 0.58 $\pm$ 0.11 & 0.79 $\pm$ 0.04& 0.99 $\pm$ 0.01& 0.15 $\pm$ 0.15 & 0.57 $\pm$ 0.07& 0.87 $\pm$\ 0.03\\
\hline
\end{tabular}
\label{tab:task3_results}
\end{table}

\subsection*{Deep Learning-based HPV Status Classification}

Finally, we explored the task of classifying HPV status using imaging and EHR data. We evaluated five deep learning models where image features from a CNN (e.g., ResNet18) were combined with EHR data using a Multi-Layer Perceptron (MLP). Using a 5-fold cross-validation, we evaluated performance with multiple metrics, including the F1-Macro score and the Area Under the ROC Curve (AUC). The ResNet18 + MLP model yielded the best classification results, achieving an AUC of 0.9379 and a weighted F1-score of 0.9277 (Table~\ref{tab:task3_results}). This best result was, however, associated with a limited balanced accuracy of 0.71 because of a low specificity (0.43) despite excellent sensitivity (0.99). This use case shows that the dataset is a valuable resource for developing machine learning models to predict molecular biomarkers from standard clinical and multimodal imaging data.



\section*{Code availability}

The HECKTOR dataset generated and analyzed in this study is publicly available at \url{https://hecktor25.grand-challenge.org/}. We respectfully request that researchers cite this paper in any work that exploits this dataset. The code for data preprocessing, along with the training and evaluation scripts for all baseline models, is available on GitHub at \url{https://github.com/BioMedIA-MBZUAI/HECKTOR2025}. The final trained model weights are also provided within the same repository.

\bibliography{sample}

\begin{thebibliography}{10}
\urlstyle{rm}
\expandafter\ifx\csname url\endcsname\relax
  \def\url#1{\texttt{#1}}\fi
\expandafter\ifx\csname urlprefix\endcsname\relax\def\urlprefix{URL }\fi
\expandafter\ifx\csname doiprefix\endcsname\relax\def\doiprefix{DOI: }\fi
\providecommand{\bibinfo}[2]{#2}
\providecommand{\eprint}[2][]{\url{#2}}

\bibitem{bhat2021head}
\bibinfo{author}{Bhat, G.~R.}, \bibinfo{author}{Hyole, R.~G.} \& \bibinfo{author}{Li, J.}
\newblock \bibinfo{title}{Head and neck cancer: Current challenges and future perspectives}.
\newblock In \emph{\bibinfo{booktitle}{Advances in cancer research}}, vol. \bibinfo{volume}{152}, \bibinfo{pages}{67--102} (\bibinfo{publisher}{Elsevier}, \bibinfo{year}{2021}).

\bibitem{Maso2016Combined}
\bibinfo{author}{Maso, L.~D.} \emph{et~al.}
\newblock \bibinfo{journal}{\bibinfo{title}{Combined effect of tobacco smoking and alcohol drinking in the risk of head and neck cancers: a re-analysis of case–control studies using bi-dimensional spline models}}.
\newblock {\emph{\JournalTitle{European Journal of Epidemiology}}} \textbf{\bibinfo{volume}{31}}, \bibinfo{pages}{385--393}, \doiprefix\url{10.1007/s10654-015-0028-3} (\bibinfo{year}{2016}).

\bibitem{Dhull2018Major}
\bibinfo{author}{Dhull, A.}, \bibinfo{author}{Atri, R.}, \bibinfo{author}{Dhankhar, R.}, \bibinfo{author}{Chauhan, A.} \& \bibinfo{author}{Kaushal, V.}
\newblock \bibinfo{journal}{\bibinfo{title}{Major risk factors in head and neck cancer: A retrospective analysis of 12-year experiences}}.
\newblock {\emph{\JournalTitle{World Journal of Oncology}}} \textbf{\bibinfo{volume}{9}}, \bibinfo{pages}{80 -- 84}, \doiprefix\url{10.14740/wjon1104w} (\bibinfo{year}{2018}).

\bibitem{Aupérin2020Epidemiology}
\bibinfo{author}{Aupérin, A.}
\newblock \bibinfo{journal}{\bibinfo{title}{Epidemiology of head and neck cancers: an update.}}
\newblock {\emph{\JournalTitle{Current Opinion in Oncology}}} \doiprefix\url{10.1097/CCO.0000000000000629} (\bibinfo{year}{2020}).

\bibitem{Gormley2022Reviewing}
\bibinfo{author}{Gormley, M.}, \bibinfo{author}{Creaney, G.}, \bibinfo{author}{Schache, A.}, \bibinfo{author}{Ingarfield, K.} \& \bibinfo{author}{Conway, D.}
\newblock \bibinfo{journal}{\bibinfo{title}{Reviewing the epidemiology of head and neck cancer: definitions, trends and risk factors}}.
\newblock {\emph{\JournalTitle{British Dental Journal}}} \textbf{\bibinfo{volume}{233}}, \bibinfo{pages}{780 -- 786}, \doiprefix\url{10.1038/s41415-022-5166-x} (\bibinfo{year}{2022}).

\bibitem{Hashibe2009Interaction}
\bibinfo{author}{Hashibe, M.} \emph{et~al.}
\newblock \bibinfo{journal}{\bibinfo{title}{Interaction between tobacco and alcohol use and the risk of head and neck cancer: pooled analysis in the international head and neck cancer epidemiology consortium.}}
\newblock {\emph{\JournalTitle{Cancer epidemiology, biomarkers \& prevention : a publication of the American Association for Cancer Research, cosponsored by the American Society of Preventive Oncology}}} \textbf{\bibinfo{volume}{18 2}}, \bibinfo{pages}{541--50}, \doiprefix\url{10.1158/1055-9965.EPI-08-0347} (\bibinfo{year}{2009}).

\bibitem{Mody2021Head}
\bibinfo{author}{Mody, M.}, \bibinfo{author}{Haddad, R.}, \bibinfo{author}{Rocco, J.}, \bibinfo{author}{Yom, S.} \& \bibinfo{author}{Saba, N.}
\newblock \bibinfo{journal}{\bibinfo{title}{Head and neck cancer}}.
\newblock {\emph{\JournalTitle{The Lancet}}} \textbf{\bibinfo{volume}{398}}, \bibinfo{pages}{2289--2299}, \doiprefix\url{10.1016/S0140-6736(21)01550-6} (\bibinfo{year}{2021}).

\bibitem{Hashibe2007Alcohol}
\bibinfo{author}{Hashibe, M.} \emph{et~al.}
\newblock \bibinfo{journal}{\bibinfo{title}{Alcohol drinking in never users of tobacco, cigarette smoking in never drinkers, and the risk of head and neck cancer: pooled analysis in the international head and neck cancer epidemiology consortium.}}
\newblock {\emph{\JournalTitle{Journal of the National Cancer Institute}}} \textbf{\bibinfo{volume}{99 10}}, \bibinfo{pages}{777--89}, \doiprefix\url{10.1093/JNCI/DJK179} (\bibinfo{year}{2007}).

\bibitem{Maasland2014Alcohol}
\bibinfo{author}{Maasland, D.~H.} \emph{et~al.}
\newblock \bibinfo{journal}{\bibinfo{title}{Alcohol consumption, cigarette smoking and the risk of subtypes of head-neck cancer: results from the netherlands cohort study}}.
\newblock {\emph{\JournalTitle{BMC Cancer}}} \textbf{\bibinfo{volume}{14}}, \bibinfo{pages}{187 -- 187}, \doiprefix\url{10.1186/1471-2407-14-187} (\bibinfo{year}{2014}).

\bibitem{Zhang2015Different}
\bibinfo{author}{Zhang, Y.} \emph{et~al.}
\newblock \bibinfo{journal}{\bibinfo{title}{Different levels in alcohol and tobacco consumption in head and neck cancer patients from 1957 to 2013}}.
\newblock {\emph{\JournalTitle{PLoS ONE}}} \textbf{\bibinfo{volume}{10}}, \doiprefix\url{10.1371/journal.pone.0124045} (\bibinfo{year}{2015}).

\bibitem{Maier1992Tobacco}
\bibinfo{author}{Maier, H.}, \bibinfo{author}{Dietz, A.}, \bibinfo{author}{Gewelke, U.}, \bibinfo{author}{Heller, W.} \& \bibinfo{author}{Weidauer, H.}
\newblock \bibinfo{journal}{\bibinfo{title}{Tobacco and alcohol and the risk of head and neck cancer}}.
\newblock {\emph{\JournalTitle{The clinical investigator}}} \textbf{\bibinfo{volume}{70}}, \bibinfo{pages}{320--327}, \doiprefix\url{10.1007/BF00184668} (\bibinfo{year}{1992}).

\bibitem{Boehm2021Harnessing}
\bibinfo{author}{Boehm, K.}, \bibinfo{author}{Khosravi, P.}, \bibinfo{author}{Vanguri, R.}, \bibinfo{author}{Gao, J.} \& \bibinfo{author}{Shah, S.~P.}
\newblock \bibinfo{journal}{\bibinfo{title}{Harnessing multimodal data integration to advance precision oncology}}.
\newblock {\emph{\JournalTitle{Nature Reviews Cancer}}} \textbf{\bibinfo{volume}{22}}, \bibinfo{pages}{114 -- 126}, \doiprefix\url{10.1038/s41568-021-00408-3} (\bibinfo{year}{2021}).

\bibitem{Chen2022Pan-cancer}
\bibinfo{author}{Chen, R.~J.} \emph{et~al.}
\newblock \bibinfo{journal}{\bibinfo{title}{Pan-cancer integrative histology-genomic analysis via multimodal deep learning.}}
\newblock {\emph{\JournalTitle{Cancer cell}}} \textbf{\bibinfo{volume}{40 8}}, \bibinfo{pages}{865--878.e6}, \doiprefix\url{10.1016/j.ccell.2022.07.004} (\bibinfo{year}{2022}).

\bibitem{Steyaert2023Multimodal}
\bibinfo{author}{Steyaert, S.} \emph{et~al.}
\newblock \bibinfo{journal}{\bibinfo{title}{Multimodal data fusion for cancer biomarker discovery with deep learning}}.
\newblock {\emph{\JournalTitle{Nature Machine Intelligence}}} \textbf{\bibinfo{volume}{5}}, \bibinfo{pages}{351--362}, \doiprefix\url{10.1038/s42256-023-00633-5} (\bibinfo{year}{2023}).

\bibitem{saeed2021ensemble}
\bibinfo{author}{Saeed, N.}, \bibinfo{author}{Al~Majzoub, R.}, \bibinfo{author}{Sobirov, I.} \& \bibinfo{author}{Yaqub, M.}
\newblock \bibinfo{title}{An ensemble approach for patient prognosis of head and neck tumor using multimodal data}.
\newblock In \emph{\bibinfo{booktitle}{3D Head and Neck Tumor Segmentation in PET/CT Challenge}}, \bibinfo{pages}{278--286} (\bibinfo{publisher}{Springer}, \bibinfo{year}{2021}).

\bibitem{saeed2022tmss}
\bibinfo{author}{Saeed, N.}, \bibinfo{author}{Sobirov, I.}, \bibinfo{author}{Al~Majzoub, R.} \& \bibinfo{author}{Yaqub, M.}
\newblock \bibinfo{title}{Tmss: An end-to-end transformer-based multimodal network for segmentation and survival prediction}.
\newblock In \emph{\bibinfo{booktitle}{International conference on medical image computing and computer-assisted intervention}}, \bibinfo{pages}{319--329} (\bibinfo{organization}{Springer}, \bibinfo{year}{2022}).

\bibitem{saeed2024survrnc}
\bibinfo{author}{Saeed, N.} \emph{et~al.}
\newblock \bibinfo{title}{Survrnc: Learning ordered representations for survival prediction using rank-n-contrast}.
\newblock In \emph{\bibinfo{booktitle}{International Conference on Medical Image Computing and Computer-Assisted Intervention}}, \bibinfo{pages}{659--669} (\bibinfo{organization}{Springer}, \bibinfo{year}{2024}).

\bibitem{hatt2009fuzzy}
\bibinfo{author}{Hatt, M.}, \bibinfo{author}{Le~Rest, C.~C.}, \bibinfo{author}{Turzo, A.}, \bibinfo{author}{Roux, C.} \& \bibinfo{author}{Visvikis, D.}
\newblock \bibinfo{journal}{\bibinfo{title}{A fuzzy locally adaptive bayesian segmentation approach for volume determination in pet}}.
\newblock {\emph{\JournalTitle{IEEE transactions on medical imaging}}} \textbf{\bibinfo{volume}{28}}, \bibinfo{pages}{881--893} (\bibinfo{year}{2009}).

\bibitem{HECKTOR2022}
\bibinfo{author}{Andrearczyk, V.}, \bibinfo{author}{Oreiller, V.}, \bibinfo{author}{Abobakr, M.}, \bibinfo{author}{Akhavanallaf, A.} \emph{et~al.}
\newblock \bibinfo{title}{Overview of the {HECKTOR} challenge at {MICCAI} 2022: Automatic head and neck tumor segmentation and outcome prediction in {PET}/{CT}}.
\newblock In \emph{\bibinfo{booktitle}{Head Neck Tumor Chall (2022)}}, \bibinfo{pages}{1--30}, \doiprefix\url{10.1007/978-3-031-27420-6_1} (\bibinfo{publisher}{Springer}, \bibinfo{year}{2023}).
\newblock \bibinfo{note}{Published online 18 March 2023; available in PMC: \url{https://www.ncbi.nlm.nih.gov/pmc/articles/PMC10171217/}}.

\bibitem{carass2020evaluating}
\bibinfo{author}{Carass, A.} \emph{et~al.}
\newblock \bibinfo{journal}{\bibinfo{title}{Evaluating white matter lesion segmentations with refined s{\o}rensen-dice analysis}}.
\newblock {\emph{\JournalTitle{Scientific reports}}} \textbf{\bibinfo{volume}{10}}, \bibinfo{pages}{8242} (\bibinfo{year}{2020}).

\bibitem{cox1972regression}
\bibinfo{author}{Cox, D.~R.}
\newblock \bibinfo{journal}{\bibinfo{title}{Regression models and life-tables}}.
\newblock {\emph{\JournalTitle{Journal of the Royal Statistical Society: Series B (Methodological)}}} \textbf{\bibinfo{volume}{34}}, \bibinfo{pages}{187--220} (\bibinfo{year}{1972}).

\bibitem{lee2018deephit}
\bibinfo{author}{Lee, C.}, \bibinfo{author}{Zame, W.~R.}, \bibinfo{author}{Yoon, J.} \& \bibinfo{author}{van~der Schaar, M.}
\newblock \bibinfo{title}{Deephit: A deep learning approach to survival analysis with competing risks}.
\newblock In \emph{\bibinfo{booktitle}{Proceedings of the AAAI Conference on Artificial Intelligence}}, vol.~\bibinfo{volume}{32}, \bibinfo{pages}{2314--2321} (\bibinfo{year}{2018}).

\bibitem{yu2011learning}
\bibinfo{author}{Yu, C.-N.}, \bibinfo{author}{Greiner, R.}, \bibinfo{author}{Lin, H.-C.} \& \bibinfo{author}{Baracos, V.}
\newblock \bibinfo{title}{Learning patient-specific cancer survival distributions as a sequence of dependent regressors}.
\newblock In \emph{\bibinfo{booktitle}{Advances in Neural Information Processing Systems}}, vol.~\bibinfo{volume}{24}, \bibinfo{pages}{1845--1853} (\bibinfo{year}{2011}).

\bibitem{rebaud2022simplicity}
\bibinfo{author}{Rebaud, L.}, \bibinfo{author}{Escobar, T.}, \bibinfo{author}{Khalid, F.}, \bibinfo{author}{Girum, K.} \& \bibinfo{author}{Buvat, I.}
\newblock \bibinfo{title}{Simplicity is all you need: out-of-the-box nnunet followed by binary-weighted radiomic model for segmentation and outcome prediction in head and neck pet/ct}.
\newblock In \emph{\bibinfo{booktitle}{3D Head and Neck Tumor Segmentation in PET/CT Challenge}}, \bibinfo{pages}{121--134} (\bibinfo{publisher}{Springer}, \bibinfo{year}{2022}).

\end{thebibliography}

\end{document}